\documentclass[10pt,twocolumn,letterpaper]{article}

\usepackage[pagenumbers]{cvpr} 

\usepackage[dvipsnames]{xcolor}

\definecolor{cvprblue}{rgb}{0.21,0.49,0.74}
\usepackage[pagebackref,breaklinks,colorlinks,citecolor=cvprblue]{hyperref}

\def\paperID{***} 
\def\confName{CVPR}
\def\confYear{2024}

\title{Side4Video: Spatial-Temporal Side Network for Memory-Efficient Image-to-Video Transfer Learning}

\author{%
Huanjin Yao$^{1,3}\textsuperscript{*}$ \qquad
Wenhao Wu$^{2,3}\textsuperscript{* \Envelope}$ \qquad
Zhiheng Li$^{1}$\\
$^1$Tsinghua University \quad $^2$The University of Sydney \quad $^3$Baidu Inc. \\ 
{\small $^{*}$ Equal Contribution \qquad \Envelope~Corresponding author} \\
}

\usepackage{bbding}
\usepackage{multirow}
\usepackage{subcaption}
\usepackage{indentfirst}
\usepackage{graphicx}
\usepackage{fontawesome}

\begin{document}
\maketitle
\begin{abstract}
Large pre-trained vision models achieve impressive success in computer vision. However, fully fine-tuning large models for downstream tasks, particularly in video understanding, can be prohibitively computationally expensive. Recent studies turn their focus towards efficient image-to-video transfer learning. Nevertheless, existing efficient fine-tuning methods lack attention to training memory usage and exploration of transferring a larger model to the video domain. In this paper, we present a novel Spatial-Temporal Side Network for memory-efficient fine-tuning large image models to video understanding, named \textbf{Side4Video}. Specifically, we introduce a lightweight spatial-temporal side network attached to the frozen vision model, which avoids the backpropagation through the heavy pre-trained model and utilizes multi-level spatial features from the original image model. Extremely memory-efficient architecture enables our method to reduce 75\% memory usage than previous adapter-based methods. In this way, we can transfer a huge ViT-E (4.4B) for video understanding tasks which is 14$\times$ larger than ViT-L (304M). Our approach achieves remarkable performance on various video datasets across unimodal and cross-modal tasks (i.e., action recognition and text-video retrieval), especially in Something-Something V1\&V2 (67.3\% \& 74.6\%), Kinetics-400 (88.6\%), MSR-VTT (52.3\%), MSVD (56.1\%) and VATEX (68.8\%).
We release our code at \url{https://github.com/HJYao00/Side4Video}.
\end{abstract}


\section{Introduction}
\label{sec:intro}

\graphicspath{{figure/}}

The success of large language models (LLMs) \cite{llama, gpt-3, T5} in understanding and generating nuanced human text has inspired similar scaling endeavors in computer vision \cite{scale_vit, vit-22B, open_clip, eva}. However, compared with image models, pre-training of large video models encounters constraints posed by large high-quality video datasets and computational resources. A prevalent method is transferring CLIP image encoder, a powerful pre-trained Vision Transformer~\cite{vit} (ViT) on web-scale image-text pairs dataset, for video tasks. Nevertheless, as the model size increases, fully fine-tuning video models is computationally expensive. The raises a critical question: How to effectively adapt large pre-trained image models, such as ViT-E~\cite{eva} with its 4.4 billion parameters, to video understanding still remains a challenge.

\begin{figure}[t]
  \centering
  \includegraphics[width=1\linewidth]{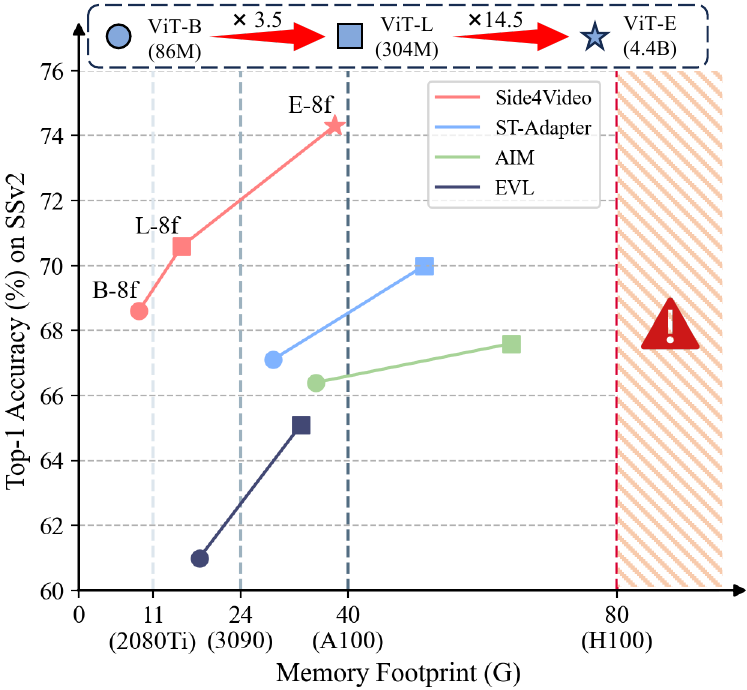}
  \vspace{-5mm}
  \caption{Comparison of GPU memory usage for training across backbones of varying parameter scales against previous efficient-training methods.}
   \label{fig:figure1}
\end{figure}

\begin{figure}
    \centering
    \includegraphics[width=0.67\linewidth]{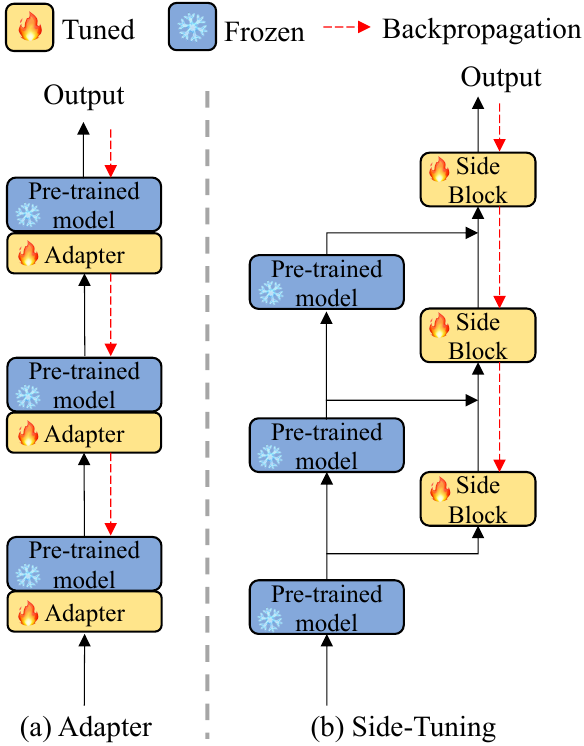}
    \vspace{-1mm}
    \caption{Illustration of Adapter-based and Side-Tuning method.}
    \label{fig:adapter_side-tuning}
\end{figure}

To accommodate the rapid expansion in model size, Parameter-Efficient Fine-Tuning (PEFT) methods \cite{lora, PEFT4nlp,lst, bitfit, prompt_tuning4nlp} which fine-tune a small part of parameters are proposed in nature language processing (NLP). Among these methods, adapter-based methods \cite{st_adapter, aim, crossadapter, uniadapter}, lightweight modules inserted into pre-trained models, are widely used for video action recognition and text-video retrieval due to efficiency and adaptability. Nevertheless, adapter-based methods require backpropagation through the frozen layers of models, which yields unnecessary memory cost, as shown in \cref{fig:adapter_side-tuning} (Left).

To further reduce memory usage, LST \cite{lst} first introduces a side network attached to the frozen pre-trained model for NLP tasks, as shown in \cref{fig:adapter_side-tuning} (Right), eliminating the need for backpropagation within the pre-trained models. A similar work \cite{side_ss} is adopted in computer vision, where a side network is used to predict mask proposals and attention bias for semantic segmentation. However, exploration of side networks in video understanding remains limited.


In this work, we introduce Side4Video, a novel and memory-efficient method for fine-tuning pre-trained image models for video understanding tasks. In addition, we explore the enhancements afforded by transferring a larger model to the task of video understanding. To be specific, we devise a spatial-temporal side network attached to frozen pre-trained models which receives multi-level spatial features from frozen ViT. Our Side4Video utilizes a divided spatial-temporal module to learn video representation which consists of temporal convolution, spatial self-attention and feed forward network in each block. Beyond simply opting for a low-dimensional side network to minimize memory usage, we investigate a variety of strategies to further conserve memory and bolster temporal reasoning capabilities, including removing [CLS] token in side network, memory-efficient temporal convolution, [CLS] token shift spatial attention. Thanks to this structure, our approach enables us to transfer ViT-E to video understanding tasks with a small amount of computational resources.


Contrary to previous PEFT methods which are applied to a single task, we evaluate our model on both unimodal and cross-modal video tasks (\ie, action recognition, and text-video retrieval) across six popular benchmarks (\ie, Something-Something (SS) V1\&V2 \cite{something}, Kinetics-400 \cite{kinetics}, MSR-VTT \cite{msrvtt}, MSVD \cite{msvd}, and VATEX \cite{vatex}).

Our contributions are summarized as follows:
\begin{itemize}
\item We introduce an innovative method for memory-efficient fine-tuning of pre-trained image models on video tasks.
\item For action recognition, our method can achieve a 75\% reduction in memory usage and a 2.2\% increase in accuracy on SSV2, surpassing the previous Video Adapter \cite{aim}. In text-video retrieval, our method achieves a 30\% memory reduction while improving the R@1 metric by 1.1 on MSR-VTT, compared to the classic CLIP4Clip~\cite{clip4clip}.
\item  To our knowledge, this is the pioneering work in efficiently transferring a large image backbone, ViT-E/14, to video understanding tasks. By scaling up the model to ViT-E/14, which is 14.5 times larger than ViT-L/14, our model delivers state-of-the-art performance on both unimodal and cross-modal video tasks.
\end{itemize}

\section{Related Work}
\label{sec:formatting}

\noindent\textbf{Large Vision Model.} The advent of ViT \cite{vit} signaled a leap forward in the pre-training of large-scale vision models, distinguished by their transferability and scalability. The CLIP model \cite{clip}, pre-trained on 400 million image-text pairs, has garnered significant interest due to its remarkable generalization capabilities and its ability to align knowledge across visual and textual domains. Building on CLIP's success, later works \cite{open_clip, eva, chineseCLIP} have expanded on the size of both datasets and models, further augmenting CLIP's representational capability. A noteworthy work is EVA-CLIP \cite{eva}, which leverages LAION-2B \cite{laion5b} consisting of 2.32 billion image-text pairs, to pre-train a 64-layer ViT-E/14 with 4.4B parameters, achieving impressive results. Yet, the efficient adaptation of such huge image models to the video domains is extremely expensive and rarely explored.


\noindent\textbf{CLIP for Video Understanding.} Due to its impressive generalization ability, CLIP is extensively expanded to action recognition \cite{atm, Xclip, uniformerv2, Text4vis, bike} and text-video retrieval \cite{wu2023cap4video,clip4clip, drl, ts2_net, uatvr, clip2video, stan}. However, these methods typically require fully fine-tuning the whole model, which is computationally intensive. To mitigate these issues, recent works \cite{st_adapter, aim, crossadapter, uniadapter, dist, clip-prompting} extend the PEFT methods \cite{lora, PEFT4nlp, bitfit, prompt_tuning4nlp, lst} from NLP to the video domain. For action recognition, ST-adapter \cite{st_adapter} and AIM ~\cite{aim} insert spatial-temporal adapters adapters inside the models to accommodate video data. For text-video retrieval, Cross-Model Adapter \cite{crossadapter} employs weight-sharing adapters into both video and text encoders. However, adapter-based methods lead to unnecessary backpropagation through the frozen parameters, incurring additional memory overhead.

\begin{figure*}[ht]
  \centering
  \includegraphics[width=1.0\linewidth]{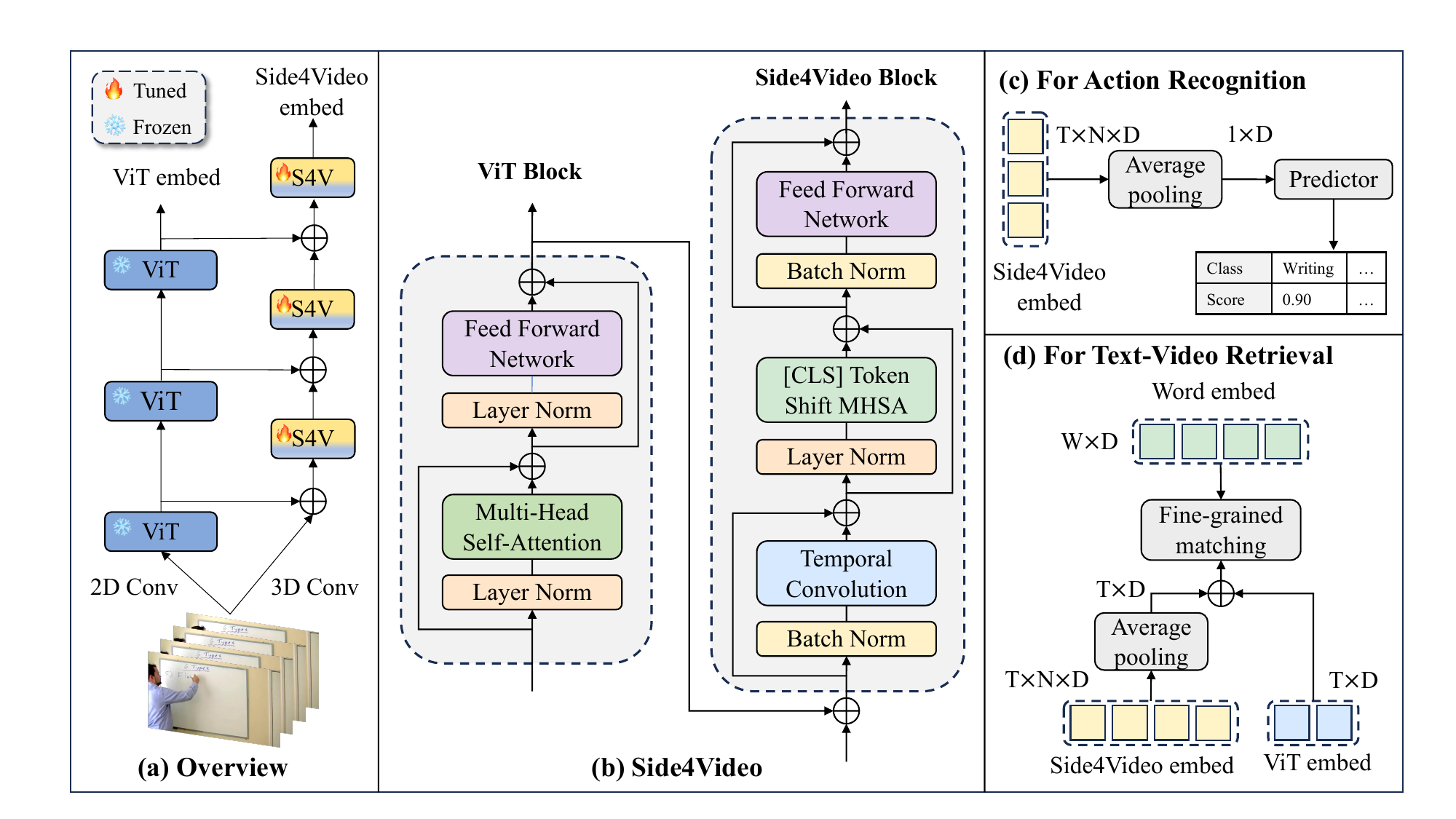}
  \caption{Illustration of our Side4Video for video understanding. (a) An overview of our Side4Video video framework. (b) The details of our Side4Video block. (c) Application of Side4Video in action recognition, and (d) its use in text-video retrieval.}
   \label{fig:network}
\end{figure*}


\noindent\textbf{Side-Tuning.} At first, Side-Tuning \cite{side-tuning} is proposed to solve the forgetfulness in incremental learning. As model size expand, fine-tuning of large models become constrained by available computational resources. LST \cite{lst} first focuses on memory reduction by implementing a lightweight transformer attached to pre-trained models for NLP tasks and SAN \cite{side_ss} leverages this technique to image semantic segmentation. These methods focus more on same modality and implement their side network by a lightweight transformer. Our work also explores cross-modal capability of side network. Note that several works \cite{evl,dist} share the similar thoughts in video domains which avoid backpropagation through the pre-trained models. EVL \cite{evl} adopts a parallel transformer decoder to extract spatial features from frozen CLIP while DiST \cite{dist} uses an integration branch to fuse the features from spatial encoder and temporal encoder, which spatial encoder is frozen CLIP. As distinct from their approaches, we introduce a spatial-temporal side encoder to learn video representation which has better continuity and scalability. Furthermore, we successfully transfer a large model for video understanding tasks to explore the advantages brought by an increased model size.
\section{Methodology}


\subsection{Preliminary} 
\label{subsec: vit and motivation}
ViT splits an image $I \in$ $\mathbb{R}^{H\times W\times C}$ into a sequence of non-overlapping patches and then project them into the embedding space as $x_e$ = [$x_1$, $x_2$, ..., $x_N$], $x_e$ $\in$ $\mathbb{R}^{N\times D}$, where $N$ denotes the number of patches and $D$ is the hidden dimension. Subsequently, ViT prepends a learnable [CLS] token $x_0$ to the $x_e$ = [$x_0$,$x_1$,$x_2$,...,$x_N$] and adds a positional embedding $E_{pos}$ to $x_e$ as $Z_0 = x_e + E_{pos}$, where $Z_0$ is the final input being fed to a sequence of transformer blocks.

Considering $T$ frames $f_t$ of a video $V$ = [$f_1$,$f_2$,...,$f_T$], our work focuses on fine-tuning a large pre-trained ViT for video understanding in a memory-efficient way. Adding adapters inside the frozen pre-trained model causes additional backpropagation through the large frozen pre-trained model. Posterior methods such as meanP \cite{clip4clip} and SeqTransf \cite{clip4clip}, which modeling spatial-temporal features after frozen ViT avoid above situation. However, posterior structures neglect low-level features which are important to video understanding tasks. Inspired by LST \cite{lst}, we propose spatial-temporal side network which utilizes multi-level spatial features to memory-efficient transfer image models to video understanding tasks.

\subsection{Overview }
\label{subsec: overview}
We introduce Side4Video, a method that fully leverages multi-level features of ViT while avoiding backpropagation through the large pre-trained models. By freezing the pre-trained models and only updating the side network parameters, our approach significantly minimizes the memory footprint. Specifically, Side4Video is constructed as a lightweight spatial-temporal side network attached to pre-trained model, consisting of $l$ layers in $d$ dimensions. The side network is seamlessly integrated with the pre-trained model, receiving multi layer features from ViT before each side block. Each Side4Video block is composed of temporal convolution, [CLS] token shift self-attention and MLP layer, as depicted in \cref{fig:network}. Finally, the output $Z_{out} \in \mathbb{R}^{T \times D}$ from ViT's [CLS] token maintains the original zero-shot capability, while the output $s_{out} \in \mathbb{R}^{T\times N\times D}$ from side network captures comprehensive video information. We deploy Global Average Pooling (GAP) on $s_{out}$ to obtain a global video representation for action recognition, while preserving frame-level global representation to support fine-grained matching for text-video retrieval.

\subsection{Side4Video}
\label{subsec: side4video}
Our Side4Video block is composed of temporal convolution, [CLS] token shift spatial self-attention and MLP layer. Here we describe Side4Video block in detail.


\noindent\textbf{Remove [CLS] token in side network.} The [CLS] token of ViT is the global image representation. In video domain, a common practice is to average [CLS] tokens of each frame as the final video representation. However, updating a learnable token increases the memory consumption. We find that GAP on patch tokens can achieve competitive performance while introducing extra [CLS] tokens in side network increase unnecessary memory footprint. Moreover, in order to enhance temporal modelling and harmonize the input paradigm \cref{eq:fusion} of each block, we use a 3D convolution project the video frames to sequence $s_0$ without additional [CLS] token, $s_0 \in \mathbb{R}^{T\times N\times d}$. 

\noindent\textbf{Feature fusion on patch tokens.} Side4Video effectively leverages multi-level spatial features of ViT. To achieve this, we implement a linear projection $Down(\cdot)$ to convert the $D$ dimensional ViT features $Z^{l}_{out}$ to $d$ dimensional features $z^{l}_{out}$. Note that this projection function is applied to both [CLS] token and patch tokens at each layer and we only fuse ViT patch tokens features $z^{l}_{out}$ and Side4Video features $s^{l-1}_{out}$ by element-wise addition. The [CLS] token will be used in spatial self-attention. The fusion strategy is:
\begin{equation}
  z^{l}_{out} = Down(Norm(Z^{l}_{out})),
  \label{eq:important}
\end{equation}
\begin{equation}
  s_{in}^{l} = s^{l-1}_{out} + z^{l}_{out}.
  \label{eq:fusion}
\end{equation}

\noindent\textbf{Temporal module in Side4Video.} Convolution \cite{3Dconv, 3dconv2, wu2021mvfnet, wu2021dsanet} and self-attention \cite{timesformer, vivit, videoSwin} are two popular way for temporal modeling. To minimize training memory cost, we investigate the impact of 3D convolution and temporal attention to memory footprint and performance, and detail is shown in \cref{subsec: ablation study}. Although temporal attention is good at long-range modeling, temporal convolution is more memory-efficient and easy to convergence. Following MVFNet~\cite{wu2021mvfnet}, we employ depth-wise separable temporal convolutions to further reduce memory. To simplify, the process starts with a $1\times 1 \times 1$ convolution as a point-wise convolution, then the $3\times 1 \times 1$ channel-wise temporal convolution followed by the $1\times 1 \times 1$ point-wise convolution to form the depth-wise separable convolution.
We also find that 3D batch normalization \cite{bn} effectively enhance spatial-temporal modeling. We adopt batch normalization before temporal convolution and MLP layer and keep layer normalization \cite{ln} before self-attention.

\noindent\textbf{[CLS] token shift self-attention.} 
Due to frozen pre-trained [CLS] tokens contain global spatial features, we extend these works \cite{tsm, ts2_net, shift_transformer} by shifting the whole pre-trained [CLS] channels back-and-forth across adjacent frames. 
Then, we concatenate the shifted token to $K$, $V$, where $K$, $V$ is the key and value in self-attention. In this case, Side4Video learn temporal information in [CLS] tokens with negligible memory increasing.

\subsection{Side4Video for video understanding}
\label{subsec: for video understanding}
Given a video, the side network generates a video representation $s_{out} \in \mathbb{R}^{T\times N\times d}$, for which we apply Global Average Pooling (GAP) over the patch tokens to obtain the final representation. We design two GAP methods to yield the final video representations for vision-only and cross-modal tasks, respectively.

\noindent\textbf{Side4Video for action recognition.} Vision-only task requires models to pay more attention on spatial-temporal modeling to understand dynamic actions. Given that the frozen pre-trained ViT lacks temporal reasoning capabilities and Side4Video models spatial-temporal features, we obtain final video representation by performing global average pooling on the output of Side4Video:

\begin{equation}
  s_{final} = \frac{1}{T\times N}\sum_{t,n}s_{out}.
  \label{eq:action recognition}
\end{equation}

\noindent\textbf{Side4Video for text-video retrieval.} Unlike vision-only task, cross-modal task requires video and text models to learn a joint embedding space. CLIP, containing rich vision-text aligned knowledge, is widely used in text-video retrieval. Since the side network is random initialized, we leverage the powerful zero-shot capabilities of CLIP to stabilize training. Specifically, we first average over the patch tokens to obtain frame-level representations of side network. Then, we project the features back to the $D$-dim and aggregate them with [CLS] tokens from the ViT. Subsequently, we reuse the pre-trained projection layer $Proj(\cdot)$ to map the features into the joint embedding space, resulting in the final frame-level representations $s_{final}$:
\begin{equation}
  s_{final} = Proj(Up(\frac{1}{N}\sum_{n}s_{out}) + Z_{out}).
  \label{eq:retrieval}
\end{equation}
Side4Video plays a role in enhancing spatial modeling and inject temporal information. 
Finally, we employ the advanced token-wise fine-grained matching~\cite{filip,drl} instead of simple global matching to generate similarity matrix for text-video retrieval.





\section{Experiments}

\subsection{Experiment Settings}

\noindent\textbf{Datasets.} To demonstrate the effectiveness of our method, we conduct a comprehensive evaluation on two popular video understanding tasks, \ie, action recognition and text-video retrieval.
For action recognition, we employ three widely adopted benchmarks to evaluate our model, including Something-Something V1\&V2 (SSV1 and SSV2) \cite{something} and Kinetics-400 (K400) \cite{kinetics}. 
For text-video retrieval, we adopt three well-known benchmarks, including MSR-VTT \cite{msrvtt}, MSVD \cite{msvd} and VATEX \cite{vatex}.
The statistics of these datasets are provided in \emph{Supplementary Material}.


\noindent\textbf{Implementation Details.}
In this paper, we adpot OpenAI-CLIP \cite{clip} for ViT-B/16, ViT-L/14 and EVA-CLIP \cite{eva} for ViT-E/14. Following the ViT-E/14, we implement flash attention \cite{flashattention} and post normalization to maintain consistency. \cref{tab:model} presents the configuration of our model for action rcognition. By adjusting dimensions and the number of layers, our model balances memory usage with performance. For text-video retrieval, we construct 320-dimensions side networks with 12, 24, and 32 layers for ViT-B, ViT-L, and ViT-E, respectively. 
Constrained by a 40G memory limit, we only train a scale-down version of ViT-E. Although the lightweight model does not fully exploit ViT-E's capabilities, it still represents a notable advancement over the ViT-L. More details are provided in \emph{Supplementary Material}.


\subsection{Memory Comparison}
\label{subsec: memory comparison}
\cref{tab:memory} presents the training memory usage and performance comparison with existing efficient fine-tuning methods on SSV2. For a fair comparison, we measure memory footprint within the same environment (A100, 80G), using 8 frames as model input. All the models are tested with 1 spatial crop and 3 temporal clips here. Benefiting from spatial-temporal side network, Our-B/16\faLightbulbO~yields a remarkable 70\% reduction in memory consumption while simultaneously improving top-1 accuracy by 1.5\% compared to ST-Adapter-B/16. Additionally, it is worth noting that another side-tuning like method DiST~\cite{dist} tends to use more tunable parameters in contrast to adapter-based methods, \ie, 19M on DiST-B \vs 7M on ST-Adapter-B. However, Our-B/16\faLightbulbO~reduce the tunable parameters to 4M which is more parameter-efficient than ST-Adapter \cite{st_adapter} and AIM \cite{aim}. Compared with DiST, Our-B/16\faLightbulbO~and Our-L/14\faLightbulbO~save 30\% and 16\% memory compared to DiST-B/16 and DiST-L/14 while achieving comparable performance. Furthermore, our method exhibits excellent scalability. Scaling up our model by increasing $l$ and $d$, Our-B/16 and Our-L/14 achieve the highest accuracy rate of 70.2\% and 71.8\%, improving by 1.5\% and 1.0\% compared to DiST-B/16 and DiST-L/14. In addition, we also provide the memory comparison on text-video retrieval in \cref{tab: retrieval memory comparison}. Compared to CLIP4Clip~\cite{clip4clip}, our method achieves a 30\% reduction in memory usage while improving 1.1\% Recall@1.

\begin{table}
  \centering
  \scalebox{0.95}{
  \begin{tabular}{@{}l|cc|cc@{}}
    \toprule
    \multirow{2}*{Methods} & \multicolumn{2}{|c|}{ViT} & \multicolumn{2}{|c}{Side4Video} \\
    ~ & layers & dim & layers & dim\\
    \midrule
    Our-B/16 \faLightbulbO &12 & 768 & 6 & 192 \\
    Our-B/16 & 12 & 768 & 12 & 320 \\
    Our-L/14 \faLightbulbO &24 & 1024 & 12 & 320 \\ 
    Our-L/14 &24 & 1024 & 24 & 512 \\
    Our-E/14 \faLightbulbO &64 & 1792 & 32 & 576\\
    \bottomrule
  \end{tabular}}
  \caption{Model configurations. We probe the performance of our model at various scales by manipulating its dimensions and the number of layers. \faLightbulbO~denotes the lightweight version.}
  \label{tab:model}
\end{table}

\begin{table}
  \centering
  \scalebox{0.95}{
  \setlength{\tabcolsep}{3.0pt}
  \begin{tabular}{@{}lcccc@{}}
    \toprule
    Methods & GFLOPs & TP (M) & Mem (G) & SSV2 (\%)  \\
    \midrule
    \textit{ViT-B/16} \\ 
    ST-Adapter~\cite{st_adapter} & 455 & 7 & 28.8& 67.1  \\
    AIM~\cite{aim} & 624 &  14 & 35.2 & 66.4\\
    EVL~\cite{evl} & 512 & 89 & 17.9 & 61.0 \\
    DiST~\cite{dist} & 480 & 19 & 12.7 & 68.7\\
    Ours \faLightbulbO & \textbf{445} & \textbf{4} & \textbf{8.9} & 68.6\\
    Ours & 528 & 21 & 18.8 & \textbf{70.2} \\
    \midrule
    \textit{ViT-L/14} \\ 
    ST-Adapter~\cite{st_adapter} & \textbf{2062} & \textbf{20} & 51.4 & 70.0 \\
    AIM~\cite{aim} & 2877 & 50 & 64.3 & 67.6\\
    EVL~\cite{evl} & 2411 & 350 & 33.0 & 65.1 \\
    DiST~\cite{dist} & 2130 & 32 & 18.1 & 70.8 \\
    Ours \faLightbulbO & 2092 & 22 & \textbf{15.3} & 70.6 \\
    Ours & 2611 & 102 & 37.0 & \textbf{71.8} \\
    \bottomrule
  \end{tabular}}
  \caption{Memory usage and performance comparison on action recognition. The memory footprint comparison of ViT-B/16 and ViT-L/14 with a batch size of 32 and 16. ``TP'' and ``Mem'' denotes the number of tunable parameters and the training memory usage.}
  \label{tab:memory}
\end{table}

\begin{table}
  \centering
  \scalebox{0.95}{
  \begin{tabular}{@{}lcc@{}}
    \toprule
    Methods & Memory (G) & R@1 (\%) \\
    \midrule
    CLIP4Clip \cite{clip4clip} & 8.2 & 43.1 \\
    CLIP4Clip (\faLock~ViT) & 3.8 & 40.0 \\
    Ours & 5.8 & \textbf{44.2}\\
    \bottomrule
  \end{tabular}}
  \caption{Memory usage comparison on MSR-VTT text-to-video retrieval. Backbone: ViT-B/16. \faLock~denotes frozen image encoder.}
  \label{tab: retrieval memory comparison}
\end{table}

\begin{table*}[htp]
  \centering
  \scalebox{0.95}{
  \setlength{\tabcolsep}{2.5pt}
  \begin{tabular}{@{}lcccccccc@{}}
    \toprule
    Method & Backbones & Pre-train & Frames$\times$Views & TFLOPs & V1 Top-1(\%) & V1 Top-5(\%) & V2 Top-1(\%) & V2 Top-5(\%) \\
    \midrule
    \textit{Full Fine-tuning} & & & &\\
    ViViT~\cite{vivit} & L/16$\times$2 FE & IN-21K+K400 & 32$\times$3$\times$4 & 1.0$\times$12 & - & - & 65.9 & 89.9 \\
    Video Swin~\cite{videoSwin} & Swin-B & IN-21K+K400 & 32$\times$3$\times$1 & 0.32$\times$3 & - & - & 69.6 & 92.7\\
    UniFormerV2~\cite{uniformerv2} & ViT-L/14 & CLIP-400M & 32$\times$3$\times$1 & 1.73$\times$3 & 62.7 & 88.0 & 73.0 & 94.5 \\
    ATM~\cite{atm} & ViT-L/14 & CLIP-400M & 16$\times$3$\times$2 & 0.84$\times$6 & 64.0 & 88.0 & 73.5 & 93.7 \\
    ATM~\cite{atm} & ViT-L/14 & Merged-2B & 16$\times$3$\times$2 & 0.84$\times$6 & 65.6 & 88.6 & 74.6 & \textbf{94.4} \\
    \midrule
    \textit{Frozen backbone} & & & &\\
    EVL-L/14~\cite{evl} & ViT-L/14 & CLIP-400M & 32$\times$1$\times$3 & 3.21$\times$3 & - & - & 66.7 & - \\
    ST-Adapter~\cite{st_adapter} & ViT-L/14 & CLIP-400M & 32$\times$1$\times$3 & 2.75$\times$3 & - & - & 72.3 & 93.9 \\
    AIM~\cite{aim} & ViT-L/14 &  CLIP-400M & 32$\times$1$\times$3 & 3.84$\times$3 & - & - & 70.6 & 92.7\\
    DiST~\cite{dist} & ViT-L/14 & CLIP-400M & 32$\times$1$\times$3 & 2.83$\times$3 & - & - & 73.1 & 93.2 \\ 
    \midrule
    Ours & ViT-B/16 & CLIP-400M & 8$\times$3$\times$2 & 0.18$\times$6 & 59.4 & 84.8 & 70.6 & 92.5 \\
    Ours & ViT-B/16 & CLIP-400M & 16$\times$3$\times$2 & 0.36$\times$6 & 60.7 & 86.0 & 71.5 & 92.8 \\
    Ours & ViT-L/14 & CLIP-400M & 8$\times$3$\times$2 & 0.87$\times$6 & 61.0 & 86.7 & 71.9 & 93.5 \\
    Ours & ViT-L/14 & CLIP-400M & 16$\times$3$\times$2 & 1.74$\times$6 & 62.4 & 88.1 & 73.2 & 93.9 \\
    Ours & ViT-E/14 & LAION-2B & 8$\times$3$\times$2 & 7.98$\times$6 & 65.3 & 88.5 & 74.3 & 94.0 \\
    Ours & ViT-E/14 & LAION-2B & 16$\times$3$\times$2 & 15.96$\times$6 & \textbf{67.3} & \textbf{88.8} & \textbf{75.2} & 94.0 \\
    \bottomrule
  \end{tabular}}
  \caption{Comparison with SOTAs on Something-Something V1\&V2. Views = \# spatial crops $\times$ \# temporal clips.}
  \label{tab:SSv2}
\end{table*}

\begin{table*}[ht]
  \centering
  \scalebox{0.95}{
  \setlength{\tabcolsep}{10pt}
  \begin{tabular}{@{}lccccccc@{}}
    \toprule
    Method & Backbones& Pre-train & Frames $\times$ Views & TFLOPs & Top-1(\%) & Top-5(\%) \\
    \midrule
    \textit{Full Fine-tuning} & & & &\\
    ViViT~\cite{vivit} & H/14$\times$2 & IN-21K & 32$\times$3$\times$4 & 3.98$\times$12 & 84.9 & 95.8 \\
    $\text{Text4Vis}^{\dag}$~\cite{wu2023transferring} & ViT-L/14 & CLIP-400M & 32$\times$3$\times$4 & 1.66$\times$12  & 87.6 & 97.8 \\
    $\text{BIKE}^{\dag}$~\cite{bike} & ViT-L/14 & CLIP-400M & 16$\times$3$\times$4 & 0.83$\times$12 & 88.1 & 97.9 \\
    ATM~\cite{atm} & ViT-L/14 & CLIP-400M & 32$\times$3$\times$4 & 1.68$\times$12 & 88.0 & 97.6\\
    \midrule
    \textit{Frozen backbone} & & & &\\
    EVL~\cite{evl} & ViT-L/14 & CLIP-400M & 32$\times$1$\times 3$ & 2.69$\times$3 & 87.3 & - \\
    ST-Adapter~\cite{st_adapter} & ViT-L/14 & CLIP-400M & 32$\times$1$\times$3 & 2.75$\times$3 & 87.2 & 97.6 \\
    AIM~\cite{aim} & ViT-L/14 & CLIP-400M & 32$\times$1$\times$3 & 3.74$\times$3 & 87.5 & 97.7 \\
    $\text{DiST}^{\dag}$~\cite{dist} & ViT-L/14& CLIP-400M & 32$\times$1$\times$3 & 2.83$\times$3 & 88.0 & 97.9 \\
    \midrule
    Ours & ViT-B/16 & CLIP-400M & 8$\times$3$\times$4 & 0.18$\times$12 & 83.6 & 96.0 \\
    Ours & ViT-B/16 & CLIP-400M & 16$\times$3$\times$4 & 0.36$\times$12 & 83.9 & 96.3 \\
    Ours & ViT-B/16 & CLIP-400M & 32$\times$3$\times$4 & 0.72$\times$12 & 84.2 & 96.5 \\
    Ours & ViT-L/14 & CLIP-400M& 8$\times$3$\times$4 & 0.87$\times$12 & 86.6 & 97.4 \\
    Ours & ViT-L/14 & CLIP-400M & 16$\times$3$\times$4 & 1.74$\times$12 & 87.0 & 97.5 \\
    Ours & ViT-E/14 & LAION-2B & 8$\times$3$\times$4 & 7.98$\times$12 & 88.3 & 98.0 \\
    Ours & ViT-E/14 & LAION-2B & 16$\times$3$\times$4 & 15.96$\times$12 & \textbf{88.6} & \textbf{98.2} \\
    \bottomrule
  \end{tabular}}
  \caption{Comparison with SOTAs on Kinetics-400. $^\dag$ represents the method utilizes textual knowledge from text encoder of CLIP. Views = \# spatial crops $\times$ \# temporal clips.}
  \label{tab:k400}
\end{table*}

\begin{table*}
  \centering
  \scalebox{0.95}{
  \setlength{\tabcolsep}{5.4pt}
  \begin{tabular}{@{}lc|ccccc|ccccc@{}}
    \toprule
    \multirow{2}*{Method} &\multirow{2}*{Pretrain} & \multicolumn{5}{c|}{Text2Video} & \multicolumn{5}{c}{Video2Text} \\
    ~ & ~ & R@1$\uparrow$ & R@5$\uparrow$ & R@10$\uparrow$ & MdR$\downarrow$ & MnR$\downarrow$ & R@1$\uparrow$ & R@5$\uparrow$ & R@10$\uparrow$ & MdR$\downarrow$ & MnR$\downarrow$\\
    \midrule
    \textit{Full Fine-tuning} & & & & & & & & \\
    CLIP4Clip~\cite{clip4clip} & CLIP-400M &  43.1 & 70.4 & 80.8 & 2.0 & 16.2 & 43.1 & 70.5 & 81.2 & 2.0 & 12.4 \\
    CLIP2Video~\cite{clip2video} & CLIP-400M & 45.6 & 72.6 & 81.7 & 2.0 & 14.6 & 43.5 & 72.3 & 82.1 & 2.0 & 10.2 \\
    STAN~\cite{stan} & CLIP-400M & 50.0 & 75.2 & 84.1 & 1.5 & - & - & - & - & - & - \\
    Cap4Video~\cite{wu2023cap4video} & CLIP-400M & 51.4 & 75.7 & 83.9 & 1.0 & \textbf{12.4} & 49.0 & 75.2 & 85.0 & 2.0 & 8.0 \\
    \midrule
    \textit{Frozen backbone} & & & & & & & &\\
    CLIP4Clip (\faLock~ViT) &CLIP-400M & 40.0 & 67.8&78.4&2.0&17.7&41.1&68.9&78.7&2.0&12.4\\
    CLIP-Prompt~\cite{clip-prompting} & CLIP-400M & 36.7 & 64.6 & 76.8 & 2.0 & - & - & - & - & - & -\\
    CM Adapter~\cite{crossadapter} &CLIP-400M&45.4&73.3&82.3&-&12.8&46.2&73.6&83.8&-&8.6\\
    UniAdapter~\cite{uniadapter} & BLIP-129M & 50.5 & 73.9 & 81.7 &1.0& - & - & - & - & - & - \\
    \midrule
    Our B/32 &CLIP-400M&44.2&71.1&81.0&2.0&15.1&44.6&72.3&82.3&2.0&9.4\\
    Our B/16 &CLIP-400M&47.2&73.8&83.7&2.0&13.1&46.6&75.8&84.3&2.0 &7.9\\
    Our L/14 &CLIP-400M&51.4&\textbf{75.8}&\textbf{84.5}&\textbf{1.0}&12.5&50.0&77.1&85.9 &\textbf{1.5}&\textbf{7.0}\\
    Our E/14 &LAION-2B&\textbf{52.3}&75.5&84.2&1.0&12.8 & \textbf{50.4} & \textbf{77.4} & \textbf{86.0} & 1.5 & 7.1 \\
    \bottomrule
  \end{tabular}}
  \caption{Results on MSR-VTT 1K. CLIP4Clip (\faLock~ViT) is our implementation with frozen image encoder. We report the results \textbf{without} any extra tricks (\eg, DSL \cite{DSL} or QB-Norm \cite{qb_norm}) during inference.}
  \label{tab:MSR-VTT}
\end{table*}

\begin{table}
  \centering
  \scalebox{0.95}{
  \setlength{\tabcolsep}{2.0pt}
  \begin{tabular}{@{}lcccccc@{}}
    \toprule
    \multirow{2}*{Method} & \multicolumn{5}{c}{Text2Video}\\
    ~ &  R@1$\uparrow$ & R@5$\uparrow$ & R@10$\uparrow$ & MdR$\downarrow$ & MnR$\downarrow$\\
    \midrule
    \textit{Full Fine-tuning} & & & & \\
    CLIP4Clip~\cite{clip4clip} & 46.2 & 76.1 & 84.6 & 2.0 & 10.0\\
    STAN~\cite{stan} & 51.5 & 80.4 & 88.5 & 1.0 & - \\
    Cap4Video~\cite{wu2023cap4video} & 51.8 & 80.8 & 88.3 & 1.0 & 8.3 \\
    \midrule
    \textit{Frozen backbone} & & & &\\
    CLIP4Clip (\faLock~ViT) & 43.8 & 73.3 & 82.7 & 2.0 & 11.1 \\
    CM Adapter~\cite{crossadapter} &47.4&76.6& 85.0 & - & 10.2\\
    \midrule
    Our B/32 & 44.6 & 74.9 & 83.5 & 2.0 & 10.2 \\
    Our B/16 & 49.0 & 78.5& 86.7 & 2.0 & 9.1\\
    Our L/14 & 54.9 & \textbf{82.1} & \textbf{89.3}& \textbf{1.0} & \textbf{7.5}\\
    Our E/14 & \textbf{56.1} & 81.7 & 88.8 & 1.0 & 8.4\\
    \bottomrule
  \end{tabular}}
  \caption{Text-to-video retrieval results on MSVD \textbf{without} any extra tricks (\eg, DSL \cite{DSL} or QB-Norm \cite{qb_norm}) during inference.}
  \label{tab:MSVD}
\end{table}

\begin{table}
  \centering
  \scalebox{0.95}{
  \setlength{\tabcolsep}{2.0pt}
  \begin{tabular}{@{}lcccccc@{}}
    \toprule
    \multirow{2}*{Method} & \multicolumn{5}{c}{Text2Video}\\
    ~ &  R@1$\uparrow$ & R@5$\uparrow$ & R@10$\uparrow$ & MdR$\downarrow$ & MnR$\downarrow$\\
    \midrule
    \textit{Full Fine-tuning} & & & & \\
    CLIP2Video~\cite{clip2video} & 57.3 & 90.0 & 95.5 & 1.0 & 3.6 \\
    TS2-Net~\cite{ts2_net} & 59.1 & 90.0 & 95.2 & 1.0 & 3.5 \\
    Cap4Video~\cite{wu2023cap4video}  & 66.6 & 93.1 & 97.0 & 1.0 & 2.7  \\
    \midrule
    \textit{Frozen backbone} & & & &\\
    CLIP4Clip (\faLock~ViT) & 55.7 & 87.6 & 93.8 & 1.0 & 4.2 \\
    CM Adapter~\cite{crossadapter} & 59.3 & 89.8 & 95.2 & - & 3.5\\
    \midrule
    Our B/32 & 58.1 & 89.2 & 94.8 & 1.0 & 3.7 \\
    Our B/16 & 61.7 & 91.5 & 96.0 & 1.0 & 3.1 \\
    Our L/14 & 67.9 & \textbf{93.9} & \textbf{97.3} & \textbf{1.0} & \textbf{2.6} \\
    Our E/14 & \textbf{68.8} & 93.5 & 97.0 & 1.0 & 2.7 \\
    \bottomrule
  \end{tabular}}
  \caption{Text-to-video retrieval results on VATEX \textbf{without} any extra tricks (\eg, DSL \cite{DSL} or QB-Norm \cite{qb_norm}) during inference.}
  \label{tab:VATEX}
\end{table}

\begin{table*}[ht]
    \begin{subtable}[t]{0.31\linewidth}
    \centering                
    \scalebox{0.95}{
      \begin{tabular}[t]{@{}ccc@{}}
        \toprule
        Method & Backbone & Top-1 Acc. \\
        \midrule
        Top & ViT-E/14 &64.0\% \\
        Interval & ViT-E/14 &\textbf{64.7\%} \\
        \bottomrule
      \end{tabular}}
      \caption{The impact of fusion layers.}
      \label{subtab:fusion layers} 
    \end{subtable}
\hfill             
    \begin{subtable}[t]{0.31\linewidth}
 \centering
 \scalebox{0.95}{
\begin{tabular}[t]{@{}ccc@{}}
            \toprule
            Temporal & Memory & Top-1 Acc. \\
            \midrule
            Attention & 20.4G & 57.9\% \\ 
            3D Conv & 18.8G & \textbf{58.5\%} \\
            \bottomrule
      \end{tabular}}
      \caption{The impact of the temporal modules.}
      \label{subtab:temporal}
    \end{subtable}%
\hfill
    \begin{subtable}[t]{0.31\linewidth}
\centering
\scalebox{0.95}{
\begin{tabular}[t]{@{}ccc@{}}
            \toprule
            Shift & Memory & Top-1 Acc. \\
            \midrule
            \XSolidBrush & 18.8G & 57.9\% \\
            \CheckmarkBold & 18.8G & \textbf{58.5\%} \\
            \bottomrule
      \end{tabular}}
      \caption{The impact of the [CLS] token shift.}
      \label{subtab:token shift}
    \end{subtable}

\bigskip    
    \begin{subtable}[t]{0.31\linewidth}
    \centering
    \scalebox{0.95}{
    \begin{tabular}[t]{@{}ccc@{}}
            \toprule
            Method & Memory & Top-1 Acc. \\
            \midrule
            \text{ViT [CLS]} & 19.2G & 54.7\% \\ 
            \text{Extra [CLS]} & 19.3G & 53.4\% \\
            GAP & 18.8G & \textbf{58.5\%} \\
            \bottomrule
      \end{tabular}}
      \caption{The study of final representation.}
      \label{subtab:GAP}
    \end{subtable}%
\hfill
    \begin{subtable}[t]{0.31\linewidth}
    \centering
    \scalebox{0.95}{
    \begin{tabular}[t]{@{}ccc@{}}
        \toprule
        Layers & Memory & Top-1 Acc. \\
        \midrule
        4 & 8.4G & 54.9\% \\
        6 & 11.0G & 56.9\% \\
        12 & 18.8G & \textbf{58.5\%} \\
        \bottomrule
      \end{tabular}}
      \caption{Study on the number of Side4Video layers.}
      \label{subtab:layers}
    \end{subtable}
\hfill
    \begin{subtable}[t]{0.31\linewidth}
    \centering
    \scalebox{0.95}{
    \begin{tabular}[t]{@{}ccc@{}}
            \toprule
            Dim & Memory & Top-1 Acc. \\
            \midrule
            128 & 11.3G & 56.7\% \\ 
            320 & 18.8G & \textbf{58.5\%} \\
            512 & 26.5G & 57.8\% \\
            \bottomrule
      \end{tabular}}
      \caption{The impact of the dimension.}
      \label{subtab:dim}
    \end{subtable}
    \caption{Ablation studies of Side4Video on Something-Something V1. Unless otherwise specified, all models use ViT-B/16 with 8 frames under the single view protocol.}
    \label{tab:ablation study}
    \end{table*}

\subsection{Comparisons on Action Recognition}
\noindent\textbf{Results on Something-Something V1\&V2.} \cref{tab:SSv2} shows the SOTA comparison on SSV1 and SSV2. Under same size pre-training models, our method achieves competitive results compared with full fine-tuning methods on both SSV1 and SSV2. For example, Side4Video with 16 frames achieve comparable results with UniFormerV2 \cite{uniformerv2} with 32 frames (62.4\% \vs 62.7\% on SSV1, 73.2\% \vs 73.0\% on SSV2). Moreover, Side4Video surpasses all the frozen backbone methods and Our L/14 outperforms ST-Adapter and AIM by 0.9\% and 2.6\% on SSV2. Scaling up backbone to ViT-E/14, we reach the highest accuracy of 67.3\% on SSV1 and 75.2\% on SSV2. We observe an impressive performance improvement as the model size increases. 

\noindent\textbf{Results on Kinetics-400.} \cref{tab:k400} presents the performance comparison on Kinetics-400. We conduct similar results with SSv1 and SSv2. On ViT-L/14, our model with an input of 16 frames achieves comparable results with frozen backbone methods. For example, Side4Video achieves comparable performance to ST-Adapter \cite{st_adapter} (87.0\% \vs 87.2\%) and AIM \cite{aim} (87.0\% \vs 87.5\%) with less input frames (16 \vs 32). Scaling up pre-trained model to ViT-E, our model enhances accuracy by 1.6\% over ViT-L/14, attaining an accuracy of 88.6\%.

\subsection{Comparisons on Text-Video Retrieval}
\label{subsec: comparison on tvr}
Beyond action recognition, we also evaluate our model on the cross-modal text-video retrieval task. Unlike other efficient fine-tuning methods \cite{uniadapter, crossadapter, clip-prompting} tailored exclusively for text-video retrieval, our work concentrates on the video component. Consequently, we keep the ViT frozen, opting to update only the side network and the text encoder. As a baseline for comparison, we present results from CLIP4Clip (\faLock~ViT), which similarly freezes the ViT and updates the text encoder.

\noindent\textbf{Performance on MSR-VTT, MSVD, and VATEX.} As shown in \cref{tab:MSR-VTT}, using a ViT-B/32 backbone, our model achieves a 1.1\% improvement in Recall@1 while reducing memory consumption by 30\% compared to the full fine-tuning approach of CLIP4Clip \cite{clip4clip}. When considering methods with a frozen backbone, our approach surpasses the baseline—CLIP4Clip (\faLock~ViT)—by a substantial 4.2\% in Recall@1. Regarding PEFT methods, our method shows notable performance enhancements. By scaling our model up to ViT-E/14, we set new state-of-the-art results with 52.3\% on MSR-VTT, 56.1\% on MSVD, and 68.8\% on VATEX, exceeding the performance of prior SOTA Cap4Video \cite{wu2023cap4video} by margins of 0.9\%, 4.3\%, and 2.2\%, respectively.


\subsection{Ablation study} 
\label{subsec: ablation study}
As shown in \cref{tab:ablation study}, we conduct ablation studies on Something-Something V1 dataset.

\noindent\textbf{The impact of fusion layers.} With memory limitations (A100 40GB), we deploy a 32-layer Side4Video for the 64-layer ViT-E/14. Hence, we explore how fusing features at varying depths impacts performance. We evaluate two fusion strategies: top and interval. The top strategy integrates high-level features from the 32nd to 64th layer while interval method fuses multi-level features every 2 layers from the beginning, \ie the 2nd, 4th, ..., 64th layers. The results in \cref{subtab:fusion layers} reveal that interval-based method yields a 0.7\% improvement in accuracy over top-based method. These findings suggest that multi-level fusion is more beneficial for video understanding tasks. 

\noindent\textbf{The study of different temporal components.} Although self-attention specializes in long-range modeling, it is more data-hungry and memory-inefficient compared to convolution. The results in \cref{subtab:temporal} show that temporal convolution reaches higher accuracy, with an increase of 0.6\% on SSV1.

\noindent\textbf{[CLS] token shift self-attention.} The token shift technique learns features of adjacent frames without an increase in memory footprint. In ViT, the [CLS] tokens summarize the spatial information of each frame. Leveraging this, we employ [CLS] token shift to enhance the temporal reasoning capability of our model. We show the effectiveness of [CLS] token shift in \cref{subtab:token shift} which brings 0.6\% accuracy improvement with a negligible increase in memory consumption due to the number of $K$,$V$ turn to $N$+1. 

\noindent\textbf{Exploration of video representation.} We explore multiple methods to obtain the final video representation, including the use of the [CLS] token within ViT, the incorporation of an additional [CLS] token in the side network, and the application of GAP. For the first method, we concatenate the dimensionality-reduced [CLS] token to the beginning of the input sequence for the side network. Note that we cannot update [CLS] token parameters which will lead to backpropagation through the pre-trained models while extra [CLS] token obtain poor performance. According to \cref{subtab:GAP}, GAP reaches both the highest performance and the least memory footprint. The reason for this performance gap between GAP and [CLS] token may be due to the learning-rate settings as mentioned in \cite{vit}. In conclusion, we adopt GAP for the final representation.

\noindent\textbf{The impact of layers and dimension} By adjusting layers $l$ and dimension $d$, we can control the memory consumption and performance of our model. Increasing $l$ or $d$ both enhances the complexity of models and memory usage, but their effects to the model are different. Increasing $l$ enables model to utilize more diverse level features from ViT while increasing $d$ enhances modeling ability of each layer. In study of layers, we base on interval fusion strategy mentioned above for 4 layers and 6 layers. As shown in \cref{subtab:layers}, increasing the layers from 4 to 12, we can see the performance gradually boost. The model fused features from all layers achieving the highest accuracy of 58.5\%. The results in \cref{subtab:dim} show that a dimension of 320 of our model achieves the best performance. A comparative evaluation of the significance of $d$ versus $l$ reveals that our model with 4 layers at 320 dimensions outperforms 12 layers with 128 dimensions, under a comparable memory footprint. In conclusion, given equivalent memory usage constraints, we prefer a higher dimensional side network to a deeper one.


\section{Conclusion}
In this paper, our motivation is to transfer large per-trained image models to video understanding tasks. To this end, we introduce Side4Video for memory-efficient image-to-video transfer learning for video understanding. Side4Video receives multi-level features from frozen ViT that avoids backpropagation through the pre-trained models. We achieve better performance than previous efficient fine-tuning methods. Scaling up model size, we transfer a huge pre-trained model (\ie, ViT-E) to video understanding tasks and observe its notable improvement. In the era of large models, we hope our work can inspire researchers who desire to fine-tune larger models in limited resources.

{
\small
\bibliographystyle{ieeenat_fullname}
\bibliography{main}
}

\clearpage
\maketitlesupplementary

\appendix

\section{Data Efficiency}
\begin{figure}[h]
    \centering
    \includegraphics[width=0.9\linewidth]{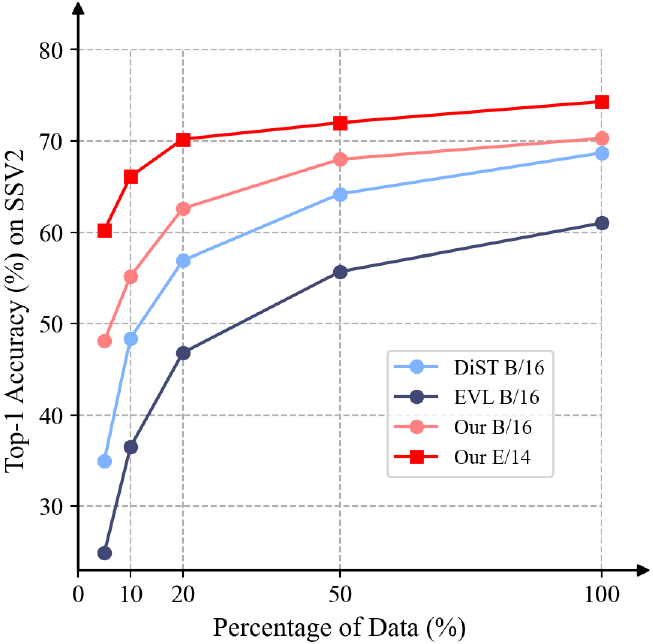}
    \caption{Comparison on data efficiency.}
    \label{fig:data efficiency}
\end{figure}

\cref{fig:data efficiency} illustrates the impact of varying training dataset sizes on the performance of our Side4Video. Our model showcases remarkable data efficiency compared with other methods. For example, with only 5\% of the Something-Something V2 dataset, Our B/16 model attains a Top-1 accuracy of 48.1\%, which is approximately 13\% higher than DiST B/16. When scaling up the backbone to ViT-E/14, our model achieves an impressive accuracy rate of 60.2\% with 5\% of the training data.

\section{Performance Gain over CLIP}
\begin{figure}[t]
    \centering
    \includegraphics[width=1.0\linewidth]{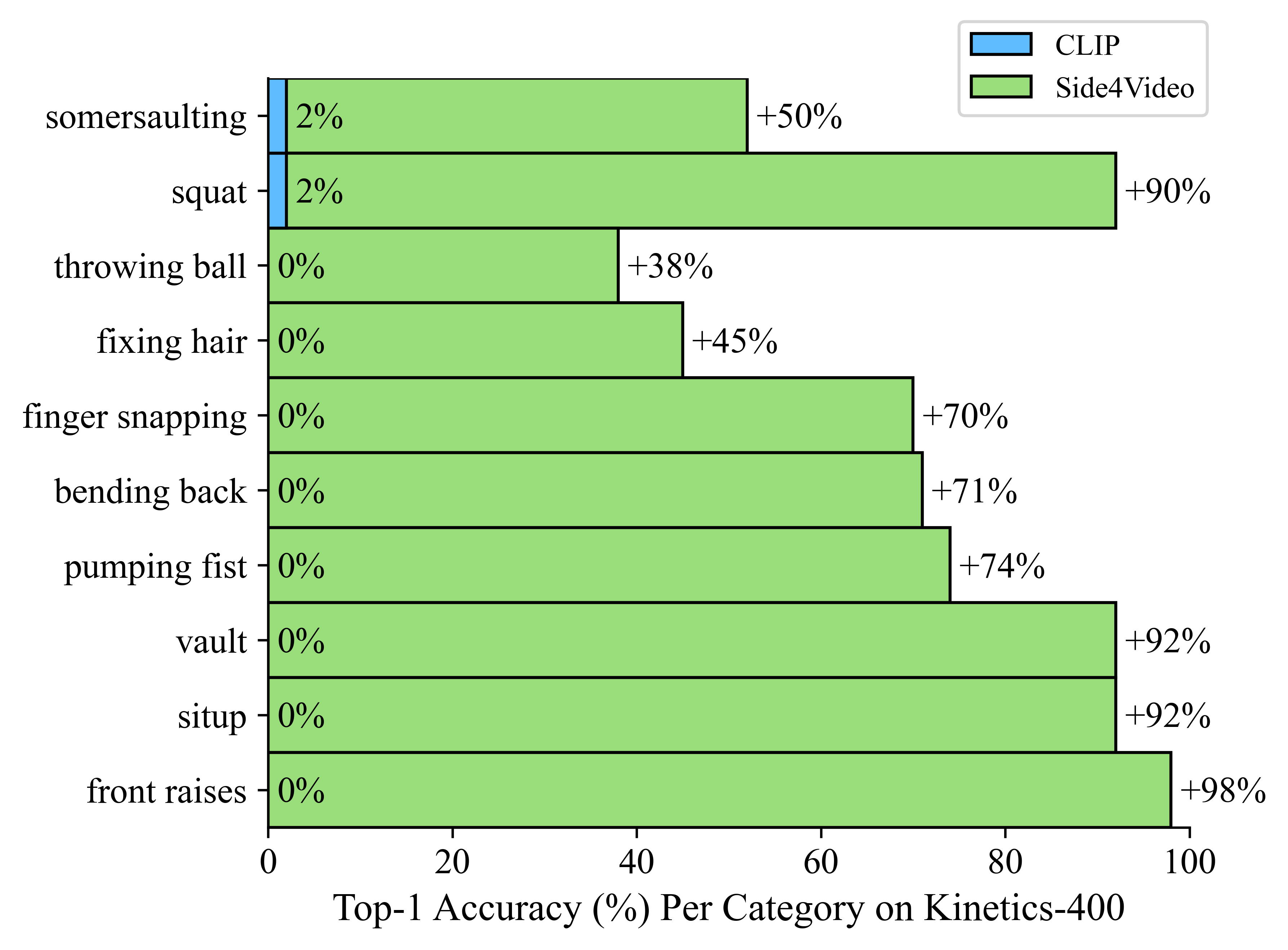}
    \caption{Comparison of CLIP and Side4Video accuracy per category on Kinetics-400. Here we only report the 10 worst category results of CLIP.}
    \label{fig: acc per category}
\end{figure}

Since our Side4Video receives multi-level spatial features from CLIP, we compare the accuracy of CLIP and Side4Video across different categories on Kinetics-400. Following \cite{wu2023transferring}, we utilize CLIP's text encoder to obtain the zero-shot classification results and \cref{fig: acc per category} reports the 10 worst classification results of CLIP. 

\begin{figure}[t]
    \centering
    \includegraphics[width=1.0\linewidth]{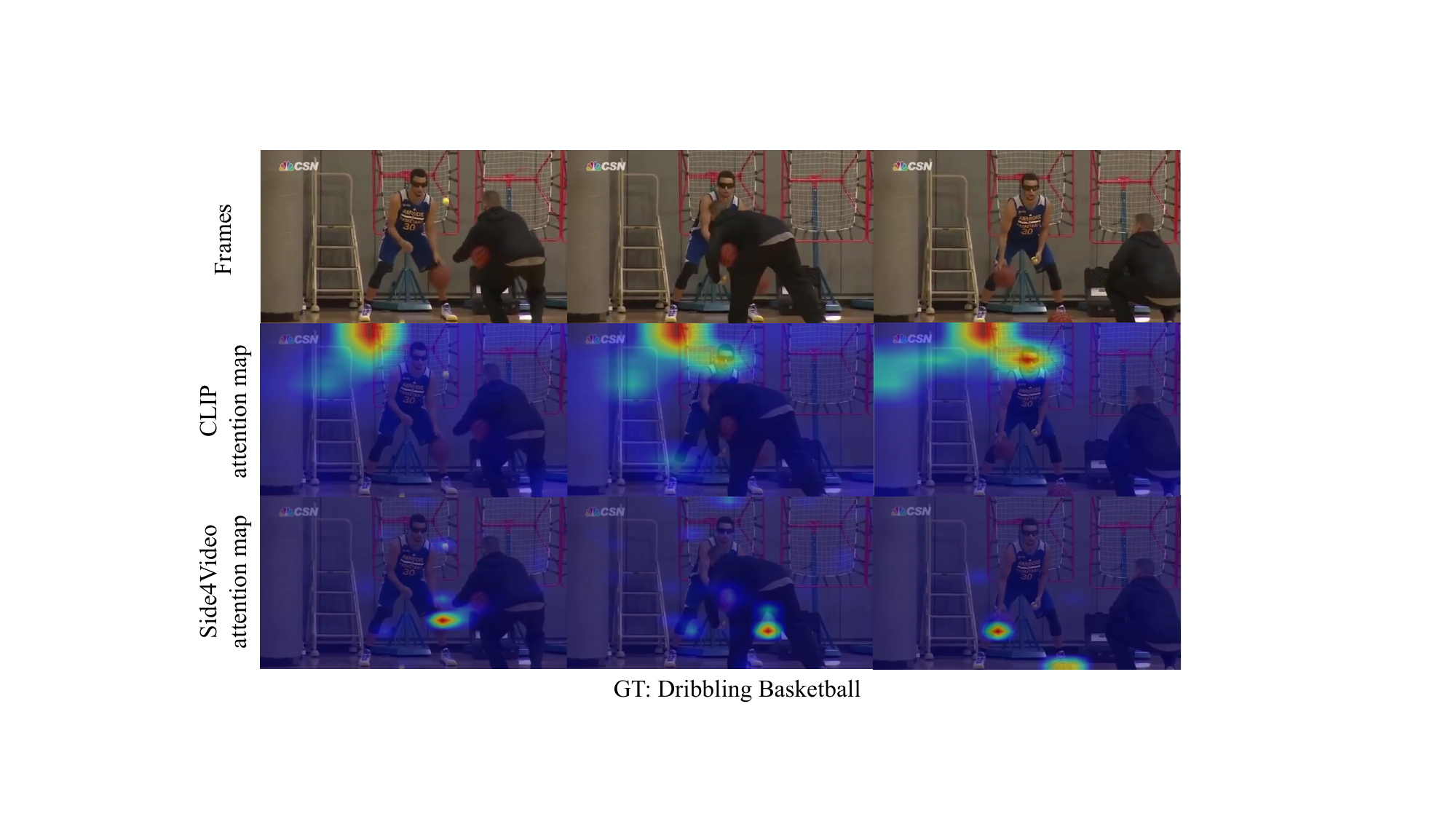}
    \caption{Visualization of Side4Video attention map. We visualize an action ``dribbling basketball" from Kinetics-400 dataset.}
    \label{fig:visualization}
\end{figure}

\section{Visualization}
In \cref{fig:visualization}, we present visualizations of the attention maps generated by CLIP and Side4Video. These illustrations demonstrate that our model more precisely concentrates on dynamically moving target objects. Significantly, as observed in the second frame, even when only a portion of the basketball is in view, our model proficiently traces the basketball's trajectory which showcases the spatial-temporal capability of our model.

\begin{table*}
  \centering
  \scalebox{0.9}{
  \setlength{\tabcolsep}{5.4pt}
  \begin{tabular}{@{}lc|ccccc|ccccc@{}}
    \toprule
    \multirow{2}*{Method} &\multirow{2}*{Pretrain} & \multicolumn{5}{c|}{Text2Video} & \multicolumn{5}{c}{Video2Text} \\
    ~ & ~ & R@1$\uparrow$ & R@5$\uparrow$ & R@10$\uparrow$ & MdR$\downarrow$ & MnR$\downarrow$ & R@1$\uparrow$ & R@5$\uparrow$ & R@10$\uparrow$ & MdR$\downarrow$ & MnR$\downarrow$\\
    \midrule
    \textit{Full Fine-tuning} & & & & & & & & \\
    CLIP4Clip~\cite{clip4clip} & CLIP-400M &46.2&76.1&84.6&2.0&10.0&56.6&79.7&84.3&1.0&7.6\\
    STAN~\cite{stan} & CLIP-400M & 51.5 & 80.4 & 88.5 & 1.0 & - &-&-&-&-&- \\
    Cap4Video~\cite{wu2023cap4video} & CLIP-400M & 51.8 & 80.8 & 88.3 & 1.0 & 8.3 &-&-&-&-&-\\
    \midrule
    \textit{Frozen backbone} & & & & & & & &\\
    CLIP4Clip (\faLock~ViT) &CLIP-400M&43.8&73.3&82.7&2.0&11.1&57.0&79.2&84.9&1.0&12.1\\
    CM Adapter~\cite{crossadapter} &CLIP-400M&47.4&76.6& 85.0 & - & 10.2&63.6&90.0&94.7&-&3.0\\
    \midrule
    Our B/32 &CLIP-400M&44.6&74.9&83.5&2.0&10.2&58.1&83.6&90.1&1.0&7.7\\
    Our B/16 &CLIP-400M&49.0&78.5&86.7&2.0&9.1&62.3&85.3&89.8&1.0&7.7\\
    Our L/14 &CLIP-400M&54.9&\textbf{82.1}&\textbf{89.3}&\textbf{1.0}&\textbf{7.5}&\textbf{71.7}&\textbf{94.4}&\textbf{97.9}&\textbf{1.0}&\textbf{2.3}\\
    Our E/14 &LAION-2B&\textbf{56.1}&81.7&88.8&1.0&8.4&65.9&90.6&95.4&1.0&2.5\\
    \bottomrule
  \end{tabular}}
  \caption{Results on MSVD. CLIP4Clip (\faLock~ViT) is our implementation with frozen image encoder. We report the results \textbf{without} any extra tricks (\eg, DSL \cite{DSL} or QB-Norm \cite{qb_norm}) during inference.}
  \label{tab:MSVD supplementary}
\end{table*}

\begin{table*}
  \centering
  \scalebox{0.9}{
  \setlength{\tabcolsep}{5.4pt}
  \begin{tabular}{@{}lc|ccccc|ccccc@{}}
    \toprule
    \multirow{2}*{Method} &\multirow{2}*{Pretrain} & \multicolumn{5}{c|}{Text2Video} & \multicolumn{5}{c}{Video2Text} \\
    ~ & ~ & R@1$\uparrow$ & R@5$\uparrow$ & R@10$\uparrow$ & MdR$\downarrow$ & MnR$\downarrow$ & R@1$\uparrow$ & R@5$\uparrow$ & R@10$\uparrow$ & MdR$\downarrow$ & MnR$\downarrow$\\
    \midrule
    \textit{Full Fine-tuning} & & & & & & & & \\
    CLIP2Video~\cite{clip2video} &CLIP-400M& 57.3 & 90.0 & 95.5 & 1.0 & 3.6 &-&-&-&-&- \\
    TS2-Net~\cite{ts2_net} &CLIP-400M& 59.1 & 90.0 & 95.2 & 1.0 & 3.5 &-&-&-&-&- \\
    Cap4Video~\cite{wu2023cap4video}  &CLIP-400M& 66.6 & 93.1 & 97.0 & 1.0 & 2.7&-&-&-&-&- \\
    \midrule
    \textit{Frozen backbone} & & & & & & & &\\
    CLIP4Clip (\faLock~ViT) &CLIP-400M& 55.7 & 87.6 & 93.8 & 1.0 & 4.2 & 76.3 & 98.0 & 99.2 & 1.0 & 1.6\\
    CM Adapter~\cite{crossadapter} &CLIP-400M& 59.3 & 89.8 & 95.2 & - & 3.5 & 74.7 & 97.2 & 99.1 & - & 1.6\\
    \midrule
    Our B/32 &CLIP-400M&58.1&89.2&94.8&1.0&3.7&77.2&97.6&99.1&1.0&1.6\\
    Our B/16 &CLIP-400M&61.7&91.5&96.0&1.0&3.1&79.0&98.0&99.5&1.0&1.5\\
    Our L/14 &CLIP-400M&67.9&\textbf{93.9}&\textbf{97.3}&\textbf{1.0}&\textbf{2.6}&\textbf{82.3}&\textbf{99.4}&\textbf{99.9}&\textbf{1.0}&\textbf{1.3}\\
    Our E/14 &LAION-2B&\textbf{68.8}&93.5&97.0&1.0&2.7&79.7&98.9&99.8&1.0&1.4\\
    \bottomrule
  \end{tabular}}
  \caption{Results on VATEX. CLIP4Clip (\faLock~ViT) is our implementation with frozen image encoder. We report the results \textbf{without} any extra tricks (\eg, DSL \cite{DSL} or QB-Norm \cite{qb_norm}) during inference.}
  \label{tab:VATEX supplementary}
\end{table*}

\begin{table*}[ht]
  \centering
  \scalebox{0.9}{
  \setlength{\tabcolsep}{8pt}
  \begin{tabular}{@{}lccccccccc@{}}
    \toprule
    \multirow{2}*{Setting} & \multicolumn{3}{c}{K400} & \multicolumn{3}{c}{SSV1} & \multicolumn{3}{c}{SSV2} \\
    ~ & ViT-B & ViT-L & ViT-E & ViT-B & ViT-L & ViT-E & ViT-B & ViT-L & ViT-E \\
    \midrule
    \textit{Optimization} \\
    Optimizer & \multicolumn{9}{c}{AdamW ($\beta_{1}$ = 0.9, $\beta_{2}$ = 0.999)} \\
    Weight decay & \multicolumn{9}{c}{0.15}\\
    Learning rate schedule & \multicolumn{9}{c}{Cosine Decay} \\
    Learning rate & 1e-3 & \multicolumn{2}{c}{1e-4} & 1e-3 & \multicolumn{2}{c}{1e-4} & 1e-3 & \multicolumn{2}{c}{1e-4} \\
    Batch size & \multicolumn{2}{c}{256} & 128 & \multicolumn{2}{c}{128} & 64 & \multicolumn{2}{c}{128} & 64 \\
    Epochs & \multicolumn{2}{c}{30} & 20 & 40 & \multicolumn{2}{c}{30} & \multicolumn{2}{c}{30} & 25\\
    Warmup epochs & \multicolumn{3}{c}{4} & \multicolumn{2}{c}{4} & 6 & \multicolumn{2}{c}{4} & 6 \\
    \midrule
    \textit{Augmentation} \\
    Training resize & \multicolumn{9}{c}{RandomResizedCrop} \\
    Random augment & - & - & \multicolumn{7}{c}{rand-m7-n4-mstd0.5-inc1}\\
    Random FLIP & \multicolumn{9}{c}{0.5}\\ 
    Label smoothing & \multicolumn{9}{c}{0.1}\\
    Repeated sampling & \multicolumn{3}{c}{1} & \multicolumn{6}{c}{2}\\
    Gray Scale & \multicolumn{3}{c}{0.2} & - & - & - & - & - & - \\
    \bottomrule
  \end{tabular}}
  \caption{Training recipe for action recognition.}
  \label{tab:implementation}
\end{table*}

\section{More Results on Text-Video Retrieval}
\cref{tab:MSVD supplementary} and \cref{tab:VATEX supplementary} present more results on MSVD and VATEX, respectively. Our method also exhibits excellent performance on video-to-text retrieval. Additionally, we observe a phenomenon where Our L/14 outperforms Our E/14 on video-to-text retrieval, which may be attributed to the pre-training data and the backbones.

\section{Additional Implementation Details}

\noindent\textbf{Dataset.}  We evaluate our model on two video understanding tasks, \ie, action recognition and text-video retrieval, to demonstrate the effectiveness of our approach.

For action recognition, we employ three widely adopted benchmarks to evaluate our model, including Something-Something V1\&V2 (SSV1 and SSV2) \cite{something} and Kinetics-400 (K400) \cite{kinetics}. Temporal-related datasets SSV1 and SSV2 contain 110K videos and 220k videos in 174 classes. Scene-based dataset K400 is a large-scale video dataset comprising 300K video clips in 400 human action classes.

For text-video retrieval, we adopt three well-known benchmarks, including MSR-VTT \cite{msrvtt}, MSVD \cite{msvd} and VATEX \cite{vatex}. MSR-VTT consists of 10k videos with 20 textual descriptions for each video. We split the dataset following \cite{multi-modal4tvr, clip4clip, howto100m}, which includes 9K videos for training and 1K videos for testing. MSVD consists of 1970 video clips with approximately 80K descriptive sentences, where train, validation, and test sets are split into 1200, 100, and 670 videos. VATEX is a relatively large dataset, containing 34,991 videos with multiple annotations. There are 25,991 videos for training, 1,500 videos for validation, and 15,500 videos for testing.

\noindent\textbf{Implementation Details.} All experiments are implemented in PyTorch. For both action recognition and text-video retrieval, we employ OpenAI-CLIP \cite{clip} for ViT-B/16, ViT-L/14, and EVA-CLIP \cite{eva} for ViT-E/14.

For action recognition, \cref{tab:implementation} presents the configuration list for Kinetics-400 \cite{kinetics} and Something-Something V1\&V2 \cite{something}.

For text-video retrieval, we freeze all the ViT encoders except the final linear projection and update Side4Video and text encoder parameters. We construct Side4Video with 12, 24, and 32 layers for ViT-B/16, ViT-L/14, and ViT-E/14, respectively. For all models, the dimensionality of the side network is set to 320. Following CLIP4Clip \cite{clip4clip}, we use a unified training setting for all the datasets (\ie, MSR-VTT \cite{msrvtt}, MSVD \cite{msvd} and VATEX \cite{vatex}). We set text length to 32 and video length to 12. We train our model with a batch size of 128 for 5 epochs with Adam ($\beta_1$ = 0.9, $\beta_2$ = 0.98) optimizer. The initial learning rate is 1e-7 for CLIP module and 1e-4 for new modules. 

\end{document}